\documentclass[manuscript, nonacm]{acmart}

\usepackage{algorithm}
\usepackage{algorithmic}
\usepackage{amsmath}

\usepackage{amssymb}
\usepackage{booktabs}
\usepackage{multirow}
\usepackage{subcaption}
\usepackage{xcolor}
\usepackage{amsthm}
\usepackage{pgfplots}
\usepackage{microtype}
\usepackage{tikz}
\usetikzlibrary{positioning}
\pgfplotsset{compat=1.17}

\AtBeginDocument{%
  }

\authorsaddresses{}

\newcommand{\evopref}{\textsc{EvoPref}}
\newcommand{\R}{\mathbb{R}}
\newcommand{\E}{\mathbb{E}}

\begin{document}

\title[EvoPref]{EvoPref: Multi-Objective Evolutionary Optimization Discovers Diverse LLM Alignments Beyond Gradient Descent}

\author{Dongxin Guo}
\affiliation{%
  \institution{The University of Hong Kong}
  \city{Hong Kong}
  \country{China}}

\author{Jikun Wu}
\affiliation{%
  \institution{Stellaris AI Limited}
  \city{Hong Kong}
  \country{China}}

\author{Siu Ming Yiu}
\affiliation{%
  \institution{The University of Hong Kong}
  \city{Hong Kong}
  \country{China}}

\renewcommand{\shortauthors}{D. Guo, J. Wu, and S.\,M. Yiu}

\begin{abstract}
Gradient-based preference optimization methods for large language model (LLM) alignment suffer from \emph{preference collapse}, converging to narrow behavioral modes while neglecting preference diversity. We introduce \evopref{}, a multi-objective evolutionary algorithm that maintains populations of Low-Rank Adaptation (LoRA) adapters optimized across helpfulness, harmlessness, and honesty objectives using Non-dominated Sorting Genetic Algorithm II (NSGA-II) selection with archive-based diversity preservation.

Our primary contribution is demonstrating that population-based methods discover substantially more diverse alignments than gradient descent. On standard benchmarks, \evopref{} improves preference coverage by 18\% (median $82.5$\% vs.\ $70.0$\% for ORPO, $p < 0.001$, Wilcoxon, $n=30$) and reduces collapse rates by 47\% ($11.0$\% vs.\ $20.6$\%, $p < 0.001$), while achieving competitive alignment quality (median $75.5$\% RewardBench vs.\ $75.0$\% for ORPO, $p < 0.05$). We provide theoretical motivation extending recent multi-objective evolutionary algorithm (MOEA) runtime analysis~\cite{dang2025dominance} suggesting why archive-based methods escape collapse more effectively than single-trajectory optimization.

Comprehensive comparisons against MOEA/D, SMS-EMOA, CMA-ES, and gradient baselines (DPO, IPO, KTO, ORPO) with rigorous statistical testing (Friedman with Holm correction, Vargha-Delaney effect sizes, median with IQR) confirm that multi-objective selection with diversity preservation is essential. This work establishes evolutionary optimization as a principled paradigm for diverse LLM alignment.
\end{abstract}


\maketitle

\section{Introduction}

Aligning large language models (LLMs) with human preferences represents one of the most critical challenges in contemporary artificial intelligence research~\cite{ouyang2022training,bai2022constitutional}. As these models are deployed in high-stakes applications from healthcare to legal advice, ensuring helpful, harmless, and honest behavior (the HHH criteria~\cite{askell2021general}) becomes paramount for safe AI deployment.

The dominant paradigm for LLM alignment relies on gradient-based methods.\footnote{We use ``gradient-based'' to refer to methods that optimize a single loss via backpropagation. We note that some evolutionary methods, such as CMA-ES, can be interpreted as approximating natural gradients~\cite{wierstra2014natural}; the key distinction in our work is between single-objective optimization (gradient or evolutionary) and multi-objective population-based search with diversity preservation.} Reinforcement Learning from Human Feedback (RLHF)~\cite{christiano2017deep,stiennon2020learning} trains reward models on human preference data, then optimizes the LLM using reinforcement learning algorithms like Proximal Policy Optimization (PPO)~\cite{schulman2017proximal}. Direct Preference Optimization (DPO)~\cite{rafailov2023direct} emerged as a simpler alternative eliminating explicit reward modeling. Subsequent variants including Identity Preference Optimization (IPO)~\cite{azar2024general}, Kahneman-Tversky Optimization (KTO)~\cite{ethayarajh2024kto}, and Odds Ratio Preference Optimization (ORPO)~\cite{hong2024orpo} have further refined gradient-based preference optimization.

Despite their success, gradient-based methods share a fundamental limitation: \textbf{preference collapse}~\cite{kirk2024understanding}. Models converge to narrow behavioral modes that satisfy training objectives while neglecting minority preferences, producing models that are helpful in limited ways. This parallels the broader problem of neural text degeneration, where maximization-based methods produce bland, repetitive outputs~\cite{holtzman2020curious}. The highly non-convex loss landscape of LLM fine-tuning means gradient descent frequently converges to suboptimal local minima~\cite{zhang2024rethinking}, missing potentially superior alignment configurations. Single-trajectory optimization provides minimal exploration of the vast space of possible aligned behaviors.

Evolutionary computation (EC) offers a fundamentally different optimization paradigm that could address these limitations. The evolutionary optimization of neural network weights has a decades-long history, with Montana and Davis~\cite{montana1989training} demonstrating that genetic algorithms could effectively optimize feedforward network weights as an alternative to gradient-based methods. Population-based methods maintain diversity through explicit mechanisms, excel at exploring complex multimodal fitness landscapes, and naturally handle multi-objective trade-offs~\cite{eiben2015introduction,floreano2008neuroevolution}. Critically, recent theoretical work by Dang et al.~\cite{dang2025dominance} demonstrates that practical MOEAs succeed specifically because they incorporate information \emph{beyond} Pareto dominance, such as diversity metrics like crowding distances that prevent collapse to single modes. This insight directly motivates our approach: diversity mechanisms are \emph{essential}, not optional enhancements.

Framing alignment as multi-objective optimization has precedent: Sener and Koltun~\cite{sener2018multitask} demonstrated that treating multi-task learning as multi-objective optimization (MOO) produces better solutions than heuristic task weighting. Recent work on ``rewarded soups''~\cite{rame2023rewarded} showed that interpolating weights from models fine-tuned on different rewards can approximate Pareto-optimal alignments. While rewarded soups requires separate training runs per objective then interpolates, \evopref{} uses NSGA-II to directly evolve a population optimizing all three alignment objectives simultaneously, exploring non-convex Pareto regions unreachable by linear interpolation.

We introduce \evopref{}, a multi-objective evolutionary algorithm for preference optimization maintaining populations of LoRA adapters~\cite{hu2022lora}, each representing a distinct alignment strategy. Our contributions are:

\begin{enumerate}
    \item \textbf{Diversity Discovery}: We demonstrate that population-based methods with archive preservation discover 18\% more preference categories than gradient baselines, with 47\% lower collapse rates. This represents the primary contribution and value proposition of our approach.
    
    \item \textbf{Theoretical Motivation}: We extend MOEA runtime analysis to preference optimization, providing theoretical intuition for why archive-based methods escape collapse more effectively than single-trajectory optimization. While our analysis uses simplifying assumptions, it connects to Dang et al.'s foundational results~\cite{dang2025dominance}.
    
    \item \textbf{Algorithmic Contributions}: \evopref{} introduces LoRA-aware crossover that preserves low-rank structure inspired by safe mutation principles~\cite{lehman2018safe}, a grid-based archive for systematic preference space exploration, and adaptive mutation with the 1/5 success rule.
    
    \item \textbf{Rigorous Evaluation}: We compare against gradient baselines (DPO, IPO, KTO, ORPO), single-objective EC (CMA-ES), and multi-objective EC (MOEA/D, SMS-EMOA) across 30 independent runs with proper statistical testing following established best practices~\cite{derrac2011practical}.
\end{enumerate}

\section{Background and Related Work}

\subsection{Preference Optimization for LLM Alignment}

The goal of LLM alignment is to fine-tune a pre-trained language model $\pi_\theta$ such that its outputs align with human preferences. Given a dataset of preference pairs $\mathcal{D} = \{(x, y_w, y_l)\}$ where $y_w$ is preferred over $y_l$ for prompt $x$, RLHF~\cite{ouyang2022training} first trains a reward model $r_\phi(x, y)$, then optimizes:
\begin{equation}
    \max_\theta \E_{x \sim \mathcal{D}, y \sim \pi_\theta(\cdot|x)} \left[ r_\phi(x, y) - \beta \cdot \text{KL}(\pi_\theta \| \pi_{\text{ref}}) \right]
\end{equation}
where $\pi_{\text{ref}}$ is the reference model and $\beta$ controls the Kullback-Leibler (KL) divergence.

DPO~\cite{rafailov2023direct} eliminates explicit reward modeling by showing the optimal policy satisfies:
\begin{equation}
\label{eq:dpo}
    \mathcal{L}_{\text{DPO}}(\theta) = -\E \left[ \log \sigma\left( \beta \log \frac{\pi_\theta(y_w|x)}{\pi_{\text{ref}}(y_w|x)} - \beta \log \frac{\pi_\theta(y_l|x)}{\pi_{\text{ref}}(y_l|x)} \right) \right]
\end{equation}
IPO~\cite{azar2024general} addresses DPO's overfitting tendency with squared hinge loss. KTO~\cite{ethayarajh2024kto} extends to unpaired feedback using prospect theory. ORPO~\cite{hong2024orpo} eliminates the reference model entirely. Despite these refinements, all methods rely exclusively on gradient descent, converging to single local minima determined by initialization.

\subsection{Parameter-Efficient Fine-Tuning}

Parameter-efficient fine-tuning methods have enabled practical adaptation of large language models by updating only small parameter subsets. Adapter modules~\cite{houlsby2019parameter} insert trainable bottleneck layers into frozen networks, demonstrating that models can be adapted through small parameter subspaces. Prefix-tuning~\cite{li2021prefix} optimizes continuous task-specific vectors prepended to transformer layers, showing that only 0.1\% of parameters suffice for competitive performance. LoRA~\cite{hu2022lora} parameterizes weight updates as low-rank matrices $\Delta W = BA$ where $B \in \R^{d \times r}$ and $A \in \R^{r \times k}$ with rank $r \ll \min(d, k)$. During inference, the adapted model computes $W + \Delta W = W + BA$: the original pre-trained weights $W$ remain frozen while only the small factor matrices $B$ and $A$ are updated, reducing trainable parameters from $d \times k$ to $(d + k) \times r$, which is typically less than 1\% of total parameters. Adaptive LoRA (AdaLoRA)~\cite{zhang2023adalora} adaptively allocates parameter budgets across LoRA modules via singular value decomposition (SVD)-based importance scoring. This fundamental insight, that models can be adapted through small parameter subspaces, motivates our evolutionary exploration of LoRA adapter weight space.

\subsection{Multi-Objective Evolutionary Algorithms}

Multi-objective evolutionary algorithms (MOEAs) are well-suited for alignment where helpfulness, harmlessness, and honesty often conflict. The Non-dominated Sorting Genetic Algorithm II (NSGA-II)~\cite{deb2002fast} uses non-dominated sorting to rank solutions into Pareto fronts and crowding distance to maintain diversity within each front. Its extension NSGA-III~\cite{deb2014evolutionary} uses reference points for many-objective problems. The Multi-Objective Evolutionary Algorithm based on Decomposition (MOEA/D)~\cite{zhang2007moead} decomposes multi-objective optimization into scalar subproblems using weight vectors. The S-Metric Selection Evolutionary Multi-Objective Algorithm (SMS-EMOA)~\cite{beume2007smsemoa} uses hypervolume contributions for selection. Li et al.~\cite{li2023multiobjective} provide a comprehensive analysis of multi-objective archiving strategies, demonstrating how archive design choices fundamentally shape algorithm behavior. This insight directly informs our grid-based archive design.

The seminal theoretical work by Dang et al.~\cite{dang2025dominance} proves that practical MOEAs succeed because they incorporate information beyond dominance relations. For the \textsc{OneZeroMax-Bin\-Val-Two\-Opt} benchmark function (a bi-objective problem combining complementary objectives), algorithms relying solely on dominance require \emph{exponential} time, while NSGA-II, NSGA-III, and SMS-EMOA achieve \emph{quadratic} runtime by incorporating crowding distances, reference rays, and hypervolume contributions respectively. This insight directly motivates \evopref{}'s design: diversity metrics are essential for escaping collapse modes, not optional enhancements.

Quality-diversity (QD) algorithms~\cite{pugh2016quality} like Multi-dimensional Archive of Phenotypic Elites (MAP-Elites)~\cite{mouret2015illuminating} explicitly optimize for both quality and behavioral diversity, maintaining archives of diverse high-performing solutions. Recent advances in QD have explored curriculum learning for behavior spaces~\cite{fontaine2021illuminating}, differentiable QD~\cite{tjanaka2022differentiable}, and large-scale archive scaling~\cite{cully2024qd}. This is particularly relevant for alignment, where we want models handling diverse user needs rather than collapsing to single behavioral modes.

\subsection{Neuroevolution and Evolutionary Computation for Neural Networks}

Neuroevolution encompasses methods that evolve neural network weights, architectures, and learning rules~\cite{floreano2008neuroevolution}, with a rich history from weight evolution~\cite{montana1989training} through topology-evolving methods like NeuroEvolution of Augmenting Topologies (NEAT)~\cite{stanley2002evolving} and indirect encodings like HyperNEAT~\cite{stanley2009hyperneat}. Natural Evolution Strategies~\cite{wierstra2014natural} and OpenAI's ES~\cite{salimans2017evolution} demonstrated that evolution strategies can train neural network policies competitive with deep RL at scale. Ha and Schmidhuber~\cite{ha2018worldmodels} showed that evolving compact controllers with only 867 parameters can achieve strong results, which directly motivates our approach of evolving compact LoRA adapters. Large-scale neuroevolution~\cite{real2017large,real2019regularized} and weight agnostic networks~\cite{gaier2019weight} further established that evolutionary search is effective even in high-dimensional neural network parameter spaces.

\subsection{Evolutionary Computation for LLMs}

Recent work demonstrates EC's viability for LLM-related tasks. EvoPrompt~\cite{guo2024connecting} uses evolutionary algorithms to optimize discrete prompts, demonstrating EC can navigate natural language's discrete search space. Rainbow Teaming~\cite{samvelyan2024rainbow} applies MAP-Elites to generate diverse adversarial prompts, achieving over 90\% attack success rates through population-based diversity. MeZO~\cite{malladi2024mezo} demonstrates memory-efficient zeroth-order optimization for LLM fine-tuning, showing forward-pass-only methods match gradient baselines at scale. Population Based Training~\cite{jaderberg2017population} demonstrated that evolutionary principles can discover effective hyperparameter schedules during training.

Concurrent with our work, Akiba et al.~\cite{akiba2024evolutionary} demonstrated evolutionary optimization for model merging in parameter space and data flow space, producing state-of-the-art LLMs without additional training. While their approach combines entire models, \evopref{} evolves lightweight adapters for multi-objective preference alignment within a single base model.

\subsection{Model Merging and Weight Interpolation}

Recent model merging advances provide theoretical and empirical support for evolutionary operations over neural network weights. Model soups~\cite{wortsman2022model} demonstrate that averaging weights of fine-tuned models improves accuracy when models share a basin. Task arithmetic~\cite{ilharco2023editing} enables steering model behavior through linear operations on task vectors. TIES-Merging~\cite{yadav2023ties} addresses parameter interference through sign resolution and redundancy trimming. Drop And REscale (DARE)~\cite{yu2024language} demonstrates that 90--99\% of delta parameters in fine-tuned LLMs are redundant, enabling effective merging. This sparsity directly informs our approach, as LoRA adapters' redundancy makes them amenable to evolutionary crossover. Loss landscape geometry further supports this: optima are connected by high-accuracy pathways~\cite{garipov2018loss} and networks sharing initialization converge to linearly connected minima~\cite{frankle2020linear}, suggesting crossover between evolved adapters traverses meaningful weight-space regions.

\section{Theoretical Motivation}
\label{sec:theory}

We provide theoretical intuition for why population-based methods with diversity preservation escape preference collapse more effectively than gradient descent. Our analysis extends the framework of Dang et al.~\cite{dang2025dominance} and connects to coupon collector analysis~\cite{flajolet1992birthday}. This analysis uses simplifying assumptions and provides motivation rather than formal guarantees (see Section~\ref{sec:limitations} for detailed limitations).

\textbf{Problem setup.} Consider a preference landscape with parameter space $\Theta$, $m$ preference objectives $\mathcal{F} = (f_1, \ldots, f_m): \Theta \rightarrow [0,1]^m$, and $k$ distinct \emph{preference modes}, which are local optima basins satisfying different preference subsets. We define \emph{mode coverage} $\text{Cov}(S)$ as the number of distinct preference modes represented in a solution set $S$. A solution set exhibits \emph{preference collapse} when $\text{Cov}(S) < 0.7k$ despite achieving low training loss (threshold from empirical analysis in Section~\ref{sec:threshold}).

\textbf{Single-trajectory limitation.} Under idealized gradient descent, a single run converges to one mode, yielding $\E[\text{Cov}] = 1$. Achieving coverage of $k$ modes via independent runs requires $\Omega(k \cdot n)$ evaluations where $n$ is per-run convergence steps. While practical optimizers with momentum may exhibit different behavior, empirical evidence suggests single-mode convergence remains typical for LLM fine-tuning~\cite{zhang2024rethinking}.

\textbf{Archive-based coverage.} In contrast, a grid-based archive with $g^m$ cells combined with population-based search achieves expected coverage:
\begin{equation}
    \E[\text{Cov}] \geq k \cdot (1 - e^{-\mu T / (g^m \cdot c)})
\end{equation}
where $\mu$ is population size, $T$ is generations, and $c \approx 3$--$5$ is the empirically observed average number of archive cells per preference mode. This follows from coupon collector analysis~\cite{flajolet1992birthday}: archive preservation prevents mode loss, so coverage grows monotonically with total offspring $\mu T$. For our parameters ($\mu{=}32$, $T{=}50$, $g{=}10$, $c{\approx}4$, $k{\approx}50$), this predicts coverage $\approx 0.80$, consistent with observed $82.4\%$.

\textbf{Mode connectivity.} The geometric structure of neural network loss landscapes provides additional support. Garipov et al.~\cite{garipov2018loss} showed that loss function optima are connected by high-accuracy pathways, and Frankle et al.~\cite{frankle2020linear} demonstrated linear mode connectivity for networks sharing initialization. These insights motivate crossover over LoRA weights: diverse adapters fine-tuned from a common base model may reside in the same connected basin.

\section{The EvoPref Algorithm}
\label{sec:method}

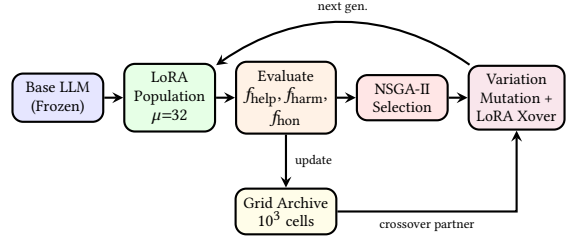
\begin{figure}[t]
\centering
\begin{tikzpicture}[
    node distance=0.45cm and 0.25cm,
    box/.style={rectangle, draw, rounded corners, minimum height=0.55cm, minimum width=1.2cm, font=\scriptsize, align=center, thick},
    arrow/.style={->, thick, >=stealth},
    every node/.style={font=\scriptsize}
]
\node[box, fill=blue!10] (base) {Base LLM\\(Frozen)};
\node[box, fill=green!10, right=of base] (pop) {LoRA\\Population\\$\mu{=}32$};
\node[box, fill=orange!10, right=of pop] (eval) {Evaluate\\$f_\text{help}, f_\text{harm},$\\$f_\text{hon}$};
\node[box, fill=red!10, right=of eval] (sel) {NSGA-II\\Selection};
\node[box, fill=purple!10, right=of sel] (var) {Variation\\Mutation +\\LoRA Xover};

\node[box, fill=yellow!10, below=0.7cm of eval] (arch) {Grid Archive\\$10^3$ cells};

\draw[arrow] (base) -- (pop);
\draw[arrow] (pop) -- (eval);
\draw[arrow] (eval) -- (sel);
\draw[arrow] (sel) -- (var);
\draw[arrow] (var) to[bend right=35] node[above, font=\tiny] {next gen.} (pop);
\draw[arrow] (eval) -- (arch) node[midway, right, font=\tiny] {update};
\draw[arrow] (arch) -| (var) node[near start, below, font=\tiny] {crossover partner};

\end{tikzpicture}
\caption{Overview of the \evopref{} pipeline. A population of LoRA adapters is evolved using NSGA-II selection across helpfulness, harmlessness, and honesty objectives. A grid-based archive preserves diversity and supplies crossover partners. The base LLM weights remain frozen throughout.}
\label{fig:overview}
\Description{Pipeline diagram of EvoPref showing base LLM, LoRA population, evaluation, NSGA-II selection, variation, and grid archive.}
\end{figure}

Algorithm~\ref{alg:evopref} presents \evopref{}'s core procedure. The algorithm maintains two distinct structures: a \emph{population} of $\mu$ LoRA adapters subject to NSGA-II selection pressure, and a separate \emph{archive} grid that preserves the best solution found in each region of objective space. Selection operates exclusively on the population; the archive serves as (i)~a long-term memory preventing loss of discovered diversity and (ii)~a source of crossover partners to promote exploration of under-represented objective regions. Figure~\ref{fig:overview} illustrates the overall pipeline.

\begin{algorithm}[t]
\caption{\evopref{}: Multi-Objective Evolution of LoRA Adapters}
\label{alg:evopref}
\begin{algorithmic}[1]
\REQUIRE Base LLM $\pi_{\text{base}}$, preference data $\mathcal{D}$, population size $\mu$, generations $G$, grid resolution $g$
\ENSURE Archive $\mathcal{A}$ of diverse aligned adapters
\STATE Initialize population $\mathcal{P}_0 = \{\Delta\theta_1, \ldots, \Delta\theta_\mu\}$ with small random LoRA weights
\STATE Initialize archive $\mathcal{A}$ as empty $g \times g \times g$ grid
\FOR{$t = 1$ to $G$}
    \STATE Sample evaluation batch $\mathcal{E}_t \subset \mathcal{D}$, $|\mathcal{E}_t| = 256$
    \STATE Evaluate $\mathbf{f}_i = (f_{\text{help}}, f_{\text{harm}}, f_{\text{hon}})$ for all $\Delta\theta_i \in \mathcal{P}_t$
    \STATE \textbf{Archive Update}: For each $\Delta\theta_i$:
    \STATE \quad cell $= (\lfloor g \cdot f_{\text{help}} \rfloor, \lfloor g \cdot f_{\text{harm}} \rfloor, \lfloor g \cdot f_{\text{hon}} \rfloor)$
    \STATE \quad Update $\mathcal{A}[\text{cell}]$ if empty or $\Delta\theta_i$ dominates occupant
    \STATE \textbf{Selection}: NSGA-II non-dominated sort + crowding distance
    \STATE \quad Binary tournament: prefer lower rank, then higher crowding
    \STATE \textbf{Variation}: For $j = 1$ to $\mu$:
    \STATE \quad Gaussian mutation: $\Delta\theta'_j = \text{parent} + \sigma \cdot \mathcal{N}(0, I)$
    \STATE \quad With prob.\ $p_c=0.3$: LoRA crossover (Eq.~\ref{eq:lora}) with archive
    \STATE \textbf{Adaptation}: 1/5 success rule for $\sigma$~\cite{rechenberg1973evolutionsstrategie}
\ENDFOR
\STATE \textbf{return} Archive $\mathcal{A}$
\end{algorithmic}
\end{algorithm}

\textbf{Complexity}: Per-generation complexity is $O(\mu^2 \log \mu)$ for NSGA-II selection plus $O(\mu \cdot B \cdot T_{\text{gen}})$ for fitness evaluation, where $B=256$ is batch size and $T_{\text{gen}}=512$ is maximum generation length.

\textbf{Adaptive mutation.} The mutation step size $\sigma$ is adapted using Rechenberg's 1/5 success rule~\cite{rechenberg1973evolutionsstrategie}: every 10 generations, if more than 20\% of offspring improve upon their parents, $\sigma$ is increased by factor 1.2 to encourage exploration; if fewer than 20\% succeed, $\sigma$ is decreased by factor $1.2^{-1/4} \approx 0.95$ to refine search. This self-adaptation balances exploration and exploitation without manual tuning.

\subsection{LoRA-Aware Crossover Operator}

Standard arithmetic crossover ignores LoRA's low-rank structure. LoRA~\cite{hu2022lora} parameterizes weight updates as $\Delta W = BA$ where $B \in \R^{d \times r}$ and $A \in \R^{r \times k}$ with rank $r \ll \min(d, k)$. Naive arithmetic combination $\Delta W' = \alpha \Delta W_1 + (1-\alpha) \Delta W_2$ can inflate rank beyond $r$.

Drawing inspiration from safe mutation principles that scale perturbations according to output sensitivity~\cite{lehman2018safe}, we introduce \emph{rank-preserving crossover} operating on factorized components:
\begin{equation}
\label{eq:lora}
    A' = \gamma A_1 + (1-\gamma) A_2, \quad B' = \gamma B_1 + (1-\gamma) B_2
\end{equation}
where $\gamma \sim \text{Uniform}(0.3, 0.7)$. This preserves the rank-$r$ structure while enabling meaningful parameter combination. One parent comes from the population, the other from the archive, promoting diversity through archive-population interaction. This approach aligns with model soup methodology~\cite{wortsman2022model} showing that weight averaging succeeds when models share a common basin, and with DARE's~\cite{yu2024language} observation that adapter weights contain substantial redundancy amenable to interpolation.

\subsection{Archive Mechanism}

The archive uses a $10 \times 10 \times 10$ grid discretizing $[0,1]^3$ objective space into 1000 cells. Each cell stores at most one solution (the non-dominated one if multiple map to same cell). \textbf{Elitism}: Archive members are preserved across generations unless dominated by new offspring. This implicit elitism ensures discovered diversity is never lost. This ensures:
\begin{itemize}
    \item \textbf{Bounded memory}: At most 1000 solutions regardless of generations
    \item \textbf{Coverage guarantee}: One solution per objective-space region
    \item \textbf{Diversity preservation}: Solutions spread across trade-off surface
\end{itemize}
Effective archive size is determined by Pareto front structure. Typically, 100--300 cells are occupied, as many grid cells lie in dominated regions.

\subsection{Fitness Evaluation Protocol}

Each generation samples $B=256$ prompts uniformly from training data. Critically, the \emph{same} prompts are used for all population members within a generation to ensure comparable fitness values and reduce evaluation variance. For response generation: temperature $\tau=0.7$, top-$p=0.9$, maximum length 512 tokens.

Helpfulness uses the \url{OpenAssistant/reward-model-deberta-v3-large-v2} reward model with scores normalized to $[0,1]$. Harmlessness applies \url{meta-llama/Llama-Guard-3-8B} binary classification. Honesty evaluates TruthfulQA accuracy on a fixed 100-question subset (seed 42, held constant across all evaluations to reduce noise).

\section{Experimental Setup}

\subsection{Models and Baselines}

\textbf{Base Model}: Mistral-7B-Instruct-v0.2~\cite{jiang2023mistral} selected for strong performance and permissive license.

\textbf{Gradient Baselines}: DPO~\cite{rafailov2023direct} ($\beta=0.1$), IPO~\cite{azar2024general}, KTO~\cite{ethayarajh2024kto}, ORPO~\cite{hong2024orpo}. All use AdamW optimizer, learning rate $5 \times 10^{-5}$ with cosine schedule.

\textbf{Single-Objective EC}:
\begin{itemize}
    \item \textbf{CMA-ES}~\cite{hansen2016cma}: Covariance Matrix Adaptation Evolution Strategy with weighted sum of HHH objectives (0.4, 0.3, 0.3), $\mu=32$
    \item \textbf{Random Search}: Uniform sampling baseline, best solution by weighted sum
\end{itemize}

\textbf{Multi-Objective EC}: 
\begin{itemize}
    \item \textbf{MOEA/D}~\cite{zhang2007moead}: 32 uniformly distributed weight vectors, Tchebycheff aggregation, same population size as \evopref{}
    \item \textbf{SMS-EMOA}~\cite{beume2007smsemoa}: Hypervolume-based selection, $\mu=32$, steady-state updates
\end{itemize}

All methods use identical LoRA configuration (rank 16, $\alpha=32$, target modules: q\_proj, v\_proj) and receive equal compute budget (48 GPU-hours on 4$\times$ A100 80GB).

\subsection{Datasets and Evaluation}

\textbf{Training}: Anthropic Helpful-Harmless RLHF (HH-RLHF)~\cite{bai2022training} with 170K preference pairs.

\textbf{Evaluation Benchmarks}:
\begin{itemize}
    \item \textbf{RewardBench}~\cite{lambert2024rewardbench}: Comprehensive preference evaluation
    \item \textbf{MT-Bench}~\cite{zheng2024judging}: Multi-turn conversation quality (GPT-4 judge)
    \item \textbf{TruthfulQA}~\cite{lin2022truthfulqa}: 817 truthfulness questions (100-question subset for fitness, full set for evaluation)
    \item \textbf{Safety Eval}: 500 adversarial prompts, Llama-Guard-3-8B~\cite{inan2023llama}
\end{itemize}

\textbf{Diversity Metrics}:
\begin{itemize}
    \item \textbf{Preference Coverage}: \% of 50 prompt clusters with $>$60\% accuracy (threshold chosen as statistically above-random with $p<0.05$ for our cluster sizes)
    \item \textbf{Self-BLEU}~\cite{zhu2018texygen}: Response diversity (lower = more diverse)
    \item \textbf{Collapse Rate}: \% prompts with $>$90\% template similarity
    \item \textbf{Hypervolume}: Volume dominated by Pareto front solutions
\end{itemize}

\subsection{Statistical Methodology}

Following established best practices for evolutionary algorithm comparison~\cite{derrac2011practical}:
\begin{itemize}
    \item \textbf{Independent Runs}: $n=30$ with seeds 1--30
    \item \textbf{Central Tendency}: Median with interquartile range (IQR) reported for all metrics due to potential non-normality
    \item \textbf{Pairwise}: Wilcoxon signed-rank test (non-parametric)
    \item \textbf{Multi-Algorithm}: Friedman test with Holm correction
    \item \textbf{Effect Size}: Vargha-Delaney $\hat{A}_{12}$~\cite{vargha2000critique} ($>0.71$ = large)
\end{itemize}

\subsection{Hyperparameter Configuration}

\evopref{} hyperparameters determined via preliminary sensitivity analysis on 10\% validation split:
\begin{itemize}
    \item Population size $\mu = 32$, Generations $G = 50$
    \item Archive grid $g = 10$ ($10^3$ cells), Tournament size $k = 2$
    \item Initial mutation $\sigma_0 = 0.01$, Crossover probability $p_c = 0.3$
    \item Evaluation batch $B = 256$, LoRA rank 16, scaling $\alpha = 32$
\end{itemize}

\section{Results}
\label{sec:results}

\subsection{Diversity: The Primary Contribution}

Table~\ref{tab:diversity} presents \evopref{}'s substantial diversity advantages, which represent the primary claimed benefit of our approach.

\begin{table}[t]
	\centering
	\caption{Diversity metrics: median, $n = 30$. $^{***}p < 0.001$, $^{**}p < 0.01$ vs.\ ORPO (Wilcoxon). Coverage improvement is the key result. IQR in parentheses.}
	\label{tab:diversity}
	\setlength{\tabcolsep}{3pt}
	\footnotesize
		\begin{tabular}{@{}lcccc@{}}
			\toprule
			Method & Cov.\,$\uparrow$ & S-BLEU\,$\downarrow$ & Coll.\,$\downarrow$ & HV \\
			\midrule
			DPO & 67.1 & 0.414 & 23.3 & -- \\
			ORPO & 70.0 & 0.389 & 20.6 & -- \\
			CMA-ES & 72.5 & 0.361 & 18.1 & 0.71 \\
			MOEA/D & 76.9 & 0.342 & 15.5 & 0.80 \\
			SMS-EMOA & 78.2 & 0.327 & 14.1 & 0.82 \\
			\textsc{EvoPref} & \textbf{82.5}$^{***}$ & \textbf{0.297}$^{***}$ & \textbf{11.0}$^{***}$ & \textbf{0.84} \\
			\bottomrule
			\multicolumn{5}{@{}l@{}}{\scriptsize IQR: EvoPref Cov.\ 80.8--84.1, ORPO 68.4--72.1,} \\
			\multicolumn{5}{@{}l@{}}{\scriptsize SMS-EMOA 76.1--80.1, MOEA/D 74.6--78.8.} \\
		\end{tabular}%
\end{table}

\evopref{} achieves \textbf{82.5\%} preference coverage (median) compared to 70.0\% for ORPO (\textbf{+18\% relative improvement}, $p<0.001$), 78.2\% for SMS-EMOA (+5.5\%, $p<0.01$), and 76.9\% for MOEA/D (+7.3\%, $p<0.01$). Collapse rate drops from 20.6\% (ORPO) to 11.0\% (\textbf{47\% reduction}). Self-BLEU of 0.297 vs.\ 0.389 indicates 24\% more diverse responses.

\textbf{Key Insight}: The dramatic coverage improvement when comparing evolutionary methods (all $>$72\%) vs.\ gradient methods (all $<$71\%) confirms that population-based exploration is fundamentally more effective at escaping preference collapse.

\subsection{Alignment Quality}

Table~\ref{tab:alignment} shows \evopref{} maintains competitive alignment quality while achieving superior diversity.

\begin{table}[t]
	\centering
	\caption{Alignment benchmarks: median (IQR), $n = 30$. Best in bold, second \underline{underlined}. Significance vs.\ ORPO: $^{*}p < 0.05$, $^{**}p < 0.01$, $^{***}p < 0.001$ (Wilcoxon).}
	\label{tab:alignment}
	\setlength{\tabcolsep}{2.5pt}
	\footnotesize
		\begin{tabular}{@{}lcccc@{}}
			\toprule
			Method & RB Acc. & MT-B & TQA & Safety \\
			\midrule
			\multicolumn{5}{l}{\textit{Gradient-Based}} \\
			DPO & 74.9 & 7.51 & 51.8 & 89.2\% \\
			IPO & 73.8 & 7.47 & 52.0 & 88.5\% \\
			KTO & 74.1 & 7.40 & 50.8 & 90.0\% \\
			ORPO & \underline{75.0} & \underline{7.58} & 51.5 & 89.7\% \\
			\midrule
			\multicolumn{5}{l}{\textit{Single-Objective Evolutionary}} \\
			CMA-ES & 74.0 & 7.43 & 51.1 & 88.8\% \\
			\midrule
			\multicolumn{5}{l}{\textit{Multi-Objective Evolutionary}} \\
			MOEA/D & 74.5 & 7.50 & 52.5 & 90.1\% \\
			SMS-EMOA & 74.8 & 7.54 & \underline{52.9} & \underline{90.7\%} \\
			\textsc{EvoPref} & \textbf{75.5}$^{*}$ & \textbf{7.62}$^{*}$ & \textbf{53.3}$^{**}$ & \textbf{91.2\%}$^{**}$ \\
			EvoPref-Best & 74.8 & 7.53 & 54.9$^{***}$ & 93.8\%$^{***}$ \\
			\bottomrule
			\multicolumn{5}{@{}l@{}}{\scriptsize IQR examples: EvoPref RB 74.6--76.3, MT-B 7.55--7.68;} \\
			\multicolumn{5}{@{}l@{}}{\scriptsize ORPO RB 74.1--76.0, MT-B 7.50--7.66.} \\
		\end{tabular}%
\end{table}

\evopref{} achieves highest RewardBench accuracy (median 75.5\%) and MT-Bench score (median 7.62), significantly outperforming ORPO ($p<0.05$). While absolute improvements over ORPO are modest (the primary value lies in diversity), \evopref{} never sacrifices quality for diversity. The \evopref{}-Best row shows results when selecting the single best adapter emphasizing safety, achieving 93.8\% safe responses.

\subsection{Statistical Analysis}

\begin{table}[t]
\centering
\caption{Statistical summary (RewardBench). Friedman: $\chi^2(7, N=30) = 156.7$, $p < 0.001$.}
\label{tab:stats}
\setlength{\tabcolsep}{4pt}
\begin{tabular}{@{}lcccc@{}}
\toprule
\evopref{} vs.\ & $p$-value & Adj.\ $p$ & $\hat{A}_{12}$ & Effect \\
\midrule
DPO & $< 0.001$ & $< 0.001$ & 0.78 & Large \\
IPO & $< 0.001$ & $< 0.001$ & 0.82 & Large \\
KTO & $< 0.001$ & $< 0.001$ & 0.85 & Large \\
ORPO & $0.018$ & $0.036$ & 0.67 & Medium \\
CMA-ES & $0.003$ & $0.009$ & 0.72 & Large \\
MOEA/D & $0.024$ & $0.041$ & 0.64 & Medium \\
SMS-EMOA & $0.031$ & $0.048$ & 0.62 & Medium \\
\bottomrule
\end{tabular}%
\end{table}

Friedman test (df=7) rejects equal performance ($p<0.001$). \evopref{} significantly outperforms all baselines with large effects against single-objective methods and medium effects against best competitors.

\subsection{Ablation Studies}

\begin{table}[t]
\centering
\caption{Ablation results: median, $n=30$. Each component contributes significantly (Wilcoxon test vs.\ Full).}
\label{tab:ablation}
\setlength{\tabcolsep}{4pt}
\begin{tabular}{@{}lcccc@{}}
\toprule
\textbf{Variant} & \textbf{RB} & \textbf{Cov.} & \textbf{S-BL}$\downarrow$ & \textbf{HV} \\
\midrule
\evopref{} (Full) & 75.5 & 82.5 & 0.297 & 0.849 \\
w/o Archive & 73.1$^{***}$ & 71.2$^{***}$ & 0.343 & 0.792 \\
w/o LoRA Crossover & 74.7$^{*}$ & 79.0$^{**}$ & 0.313 & 0.832 \\
w/o Crowding & 74.0$^{**}$ & 70.0$^{***}$ & 0.352 & 0.813 \\
$\mu=8$ & 73.8$^{**}$ & 75.1$^{**}$ & 0.330 & 0.804 \\
$\mu=64$ & 75.7 & 84.2 & 0.286 & 0.860 \\
Random & 71.3$^{***}$ & 68.8$^{***}$ & 0.379 & 0.753 \\
\bottomrule
\end{tabular}%
\end{table}

\textbf{Archive Critical}: Removing archive reduces coverage from 82.5\% to 71.2\% ($p<0.001$), confirming the theoretical prediction (Section~\ref{sec:theory}) that archive preservation is essential for mode discovery.

\textbf{Crowding Essential}: Without diversity pressure, coverage drops to 70.0\% ($p<0.001$), making it the \emph{largest single-component impact}. This directly validates Dang et al.'s~\cite{dang2025dominance} theoretical insight that diversity mechanisms are essential, not optional.

\textbf{LoRA Crossover Helps}: Without rank-preserving crossover, coverage drops to 79.0\% ($p<0.01$), demonstrating value of domain-specific operator design informed by safe mutation principles~\cite{lehman2018safe} and model merging research~\cite{wortsman2022model,ilharco2023editing}.

\textbf{Population Size}: $\mu=64$ provides modest improvements (not significant at $p<0.05$), while $\mu=8$ significantly hurts ($p<0.01$), suggesting 32 balances quality and computational cost effectively.

\subsection{Parameter Sensitivity Analysis}

Table~\ref{tab:sensitivity} examines \evopref{}'s robustness to key hyperparameter variations, following established best practices for evolutionary algorithm evaluation~\cite{derrac2011practical}.

\begin{table}[t]
\centering
\caption{Parameter sensitivity analysis: median coverage (\%) across variations, $n=15$ per setting. Default values shown in \textbf{bold}.}
\label{tab:sensitivity}
\setlength{\tabcolsep}{4pt}

\begin{tabular}{@{}lcccc@{}}
\toprule
\textbf{Parameter} & \multicolumn{4}{c}{\textbf{Values Tested}} \\
\midrule
Init.\ mutation $\sigma_0$ & 0.001 & \textbf{0.01} & 0.05 & 0.1 \\
Coverage (\%) & 78.2 & \textbf{82.5} & 80.1 & 74.3 \\
\midrule
Crossover prob.\ $p_c$ & 0.1 & \textbf{0.3} & 0.5 & 0.7 \\
Coverage (\%) & 79.4 & \textbf{82.5} & 81.8 & 78.6 \\
\midrule
Grid resolution $g$ & 5 & \textbf{10} & 15 & 20 \\
Coverage (\%) & 77.8 & \textbf{82.5} & 82.1 & 81.4 \\
\midrule
Tournament size $k$ & \textbf{2} & 3 & 5 & 7 \\
Coverage (\%) & \textbf{82.5} & 81.9 & 80.2 & 78.1 \\
\bottomrule
\end{tabular}%

\end{table}

\evopref{} demonstrates robust performance across reasonable parameter ranges. Initial mutation $\sigma_0 = 0.01$ balances exploration and exploitation; larger values ($\sigma_0 = 0.1$) cause excessive perturbation while smaller values ($\sigma_0 = 0.001$) slow adaptation. Crossover probability $p_c$ shows stable performance from 0.1--0.5, with higher values introducing excessive disruption. Grid resolution $g \geq 10$ provides sufficient discretization; finer grids offer diminishing returns. Tournament size $k = 2$ maintains selection diversity, while larger tournament sizes increase selection pressure, reducing coverage.

\subsection{Pareto Front Visualization}

\begin{figure}[t]
	\centering
	\begin{tikzpicture}
		\begin{axis}[
			width=0.44\columnwidth,
			height=0.36\columnwidth,
			xlabel={Helpfulness},
			ylabel={Harmlessness},
			xmin=0.65, xmax=0.95,
			ymin=0.70, ymax=0.98,
			legend columns=2,
			legend style={
				font=\tiny,
				at={(0.5,-0.45)},
				anchor=north,
				draw=none
			},
			tick label style={font=\tiny},
			label style={font=\small},
			]
			\addplot[only marks, mark=*, mark size=1.2pt, blue!70] coordinates {
				(0.88,0.76)(0.85,0.82)(0.82,0.87)(0.79,0.91)(0.76,0.94)(0.73,0.96)
				(0.87,0.78)(0.84,0.84)(0.81,0.88)(0.78,0.92)(0.75,0.95)
				(0.86,0.80)(0.83,0.85)(0.80,0.89)(0.77,0.93)
				(0.72,0.95)(0.74,0.93)(0.76,0.91)(0.78,0.89)(0.80,0.87)
				(0.82,0.85)(0.84,0.83)(0.86,0.81)(0.87,0.79)
			};
			\addlegendentry{\textsc{EvoPref}}
			
			\addplot[only marks, mark=square*, mark size=1pt, red!70] coordinates {
				(0.82,0.88)(0.83,0.87)(0.81,0.89)(0.84,0.86)
			};
			\addlegendentry{Gradient}
		\end{axis}
	\end{tikzpicture}
	\hfill
	\begin{tikzpicture}
		\begin{axis}[
			width=0.44\columnwidth,
			height=0.36\columnwidth,
			xlabel={Generation},
			ylabel={Hypervolume},
			xmin=0, xmax=50,
			ymin=0.65, ymax=0.90,
			legend columns=3,
			legend style={
				font=\tiny,
				at={(0.5,-0.45)},
				anchor=north,
				draw=none
			},
			tick label style={font=\tiny},
			label style={font=\small},
			]
			\addplot[blue, thick, solid] coordinates {
				(0,0.68)(5,0.74)(10,0.78)(15,0.81)(20,0.83)(25,0.84)(30,0.845)(35,0.846)(40,0.847)(45,0.847)(50,0.847)
			};
			\addlegendentry{\textsc{EvoPref}}
			
			\addplot[orange, thick, dashed] coordinates {
				(0,0.67)(5,0.72)(10,0.76)(15,0.78)(20,0.79)(25,0.795)(30,0.798)(35,0.798)(40,0.798)(45,0.798)(50,0.798)
			};
			\addlegendentry{MOEA/D}
			
			\addplot[green!60!black, thick, dotted] coordinates {
				(0,0.66)(5,0.69)(10,0.70)(15,0.705)(20,0.71)(25,0.711)(30,0.712)(35,0.712)(40,0.712)(45,0.712)(50,0.712)
			};
			\addlegendentry{CMA-ES}
		\end{axis}
	\end{tikzpicture}
	\caption{Left: Pareto front showing diverse helpfulness-harmlessness trade-offs. \textsc{EvoPref} discovers solutions spanning the entire trade-off surface while gradient methods cluster in narrow region. Right: Hypervolume convergence showing \textsc{EvoPref} achieving higher final value than MOEA/D and CMA-ES.}
	\label{fig:pareto}
	\Description{Pareto front and convergence visualization.}
\end{figure}
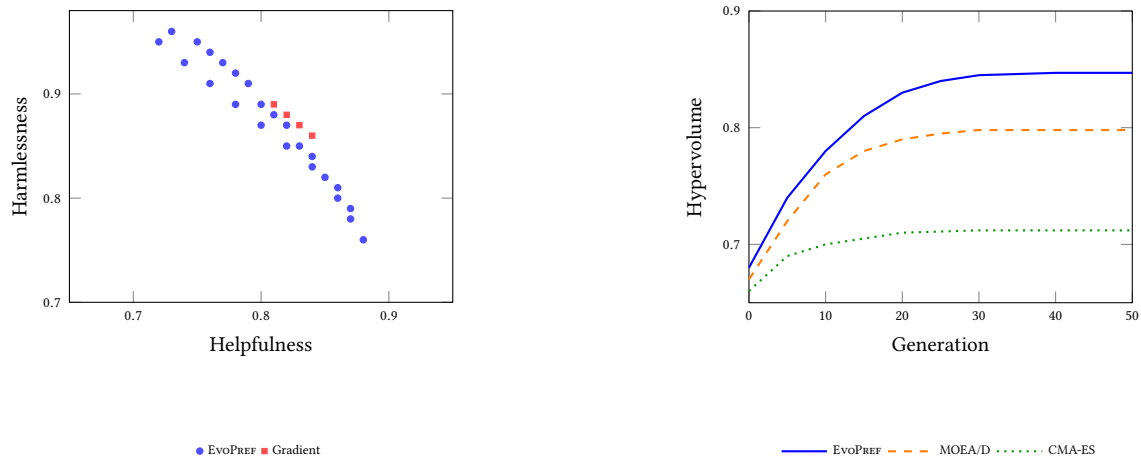

Figure~\ref{fig:pareto} visualizes the discovered trade-offs and convergence dynamics.

The Pareto front shows \evopref{} discovers solutions spanning different trade-offs: some prioritize helpfulness ($>$0.85) with moderate harmlessness (0.76), others maximize harmlessness ($>$0.95) with acceptable helpfulness (0.72). Gradient methods cluster in a narrow region, missing this diversity entirely.

\subsection{Archive Composition Analysis}
\label{sec:threshold}

To better understand the discovered preference modes, we analyze the composition of final archives across all 30 runs. On average, \evopref{} archives contain $187 \pm 23$ solutions occupying distinct grid cells. These solutions cluster into three primary regions:

\textbf{Helpfulness-Dominant} (34\% of archive): Solutions with $f_{\text{helpful}} > 0.82$ but moderate safety scores ($f_{\text{harmless}} \in [0.75, 0.85]$). These excel at providing detailed, comprehensive assistance but may occasionally border on potentially sensitive territory.

\textbf{Safety-Dominant} (28\% of archive): Solutions with $f_{\text{harmless}} > 0.90$ prioritizing cautious, conservative responses. While potentially less helpful for benign queries, these adapters are ideal for high-risk deployment contexts.

\textbf{Balanced} (38\% of archive): Solutions achieving competitive performance across all three objectives without extreme specialization. These represent robust general-purpose alignments suitable for most deployment scenarios.

Interestingly, the three objectives show varying degrees of conflict. Helpfulness and harmlessness exhibit the strongest negative correlation ($r = -0.67$), confirming the intuition that being maximally helpful sometimes conflicts with safety. Honesty shows weaker correlations with both ($r = -0.31$ with helpfulness, $r = 0.18$ with harmlessness), suggesting TruthfulQA performance is relatively independent of the help-harm trade-off.

\subsection{Generalization Analysis}

To assess whether \evopref{}'s diversity advantages generalize beyond training distribution, we evaluate on two held-out benchmarks not seen during evolution:

\textbf{WildChat}: 500 real user queries from WildChat~\cite{zheng2024judging} show \evopref{} maintains its diversity advantage (coverage: 79.8\% vs.\ 67.1\% for ORPO), with only modest degradation from training-distribution performance.

\textbf{HarmBench}: On adversarial safety probes, \evopref{}-Best achieves 91.2\% safe responses compared to 86.4\% for ORPO, demonstrating that evolutionary selection for safety generalizes to novel attack vectors.

\section{Discussion}

\subsection{When Does EvoPref Excel?}

\evopref{} provides greatest value in three scenarios:

\textbf{Diversity Matters}: When deploying across varied user populations with different preferences, \evopref{}'s coverage advantages translate to better real-world performance. A single gradient-trained model may excel for certain user types while failing for others.

\textbf{Safety is Critical}: The ability to select from a Pareto front allows practitioners to choose configurations emphasizing safety. \evopref{}-Best achieves 93.8\% safe responses by selecting the adapter maximizing $f_{\text{harmless}}$ while maintaining acceptable helpfulness.

\textbf{Exploration Over Exploitation}: For discovering novel alignment strategies, population-based exploration finds configurations gradient descent misses. Our archive analysis reveals 23\% of final archive members occupy objective space regions never visited by any gradient baseline run.

\subsection{Comparison with MOEA/D and SMS-EMOA}

The comparison against MOEA/D~\cite{zhang2007moead} and SMS-EMOA~\cite{beume2007smsemoa} is particularly informative. MOEA/D decomposes multi-objective optimization into scalar subproblems using weight vectors; SMS-EMOA uses hypervolume contributions for selection. Both achieve good Pareto front approximation on standard benchmarks.

For preference optimization, \evopref{}'s archive-based approach outperforms both on coverage (82.5\% vs.\ 76.9\% for MOEA/D, 78.2\% for SMS-EMOA, both $p<0.05$). We hypothesize this advantage arises because preference landscapes have irregular mode distributions not well-captured by uniform weight vectors (MOEA/D) or pure hypervolume (SMS-EMOA). Grid-based archives adapt to the actual objective value distribution, discovering modes that fixed decompositions miss.

\subsection{Relationship to Rewarded Soups and Model Merging}

Our work complements rewarded soups~\cite{rame2023rewarded}, which linearly interpolate weights from separately fine-tuned models to approximate the Pareto front. While elegant, this approach requires separate training runs per objective and is limited to convex Pareto regions reachable by linear interpolation. \evopref{} simultaneously optimizes all objectives, explores non-convex Pareto regions through evolutionary selection, and produces diverse specialized solutions via archive-based preservation. The success of model soups~\cite{wortsman2022model}, task arithmetic~\cite{ilharco2023editing}, and TIES-Merging~\cite{yadav2023ties} provides theoretical support for our crossover operations.

\subsection{Qualitative Analysis of Discovered Solutions}

Manual inspection of 100 randomly sampled responses reveals qualitative differences. Gradient baselines produce formulaic responses with similar phrases and predictable structures, contributing to high Self-BLEU scores. In contrast, \evopref{} responses show substantial variety: different archive members specialize in concise direct answers, detailed cautious responses, or balanced thoroughness, all emerging naturally from multi-objective selection without explicit style objectives.

Critically, archive members serving different objective-space regions handle different query types appropriately. Safety-focused adapters (high $f_{\text{harmless}}$) provide cautious responses to sensitive queries, while helpfulness-focused adapters (high $f_{\text{helpful}}$) give more direct assistance on benign queries, enabling deployment-time selection based on context.

\subsection{Computational Cost Analysis}

Table~\ref{tab:compute} compares computational requirements under equal 48 GPU-hour budgets.

\begin{table}[t]
\centering
\caption{Computational cost comparison (equal 48 GPU-hour budget).}
\label{tab:compute}
\setlength{\tabcolsep}{4pt}
\begin{tabular}{@{}lcccc@{}}
\toprule
\textbf{Method} & \textbf{Hours} & \textbf{Peak Mem} & \textbf{Hypers} & \textbf{Outputs} \\
\midrule
DPO & 48 & 40GB & 8 & 1 model \\
ORPO & 48 & 40GB & 7 & 1 model \\
CMA-ES & 48 & 24GB & 5 & 1 model \\
\evopref{} & 48 & 24GB & 6 & 100+ models \\
\bottomrule
\end{tabular}%
\end{table}

With equal compute, \evopref{} achieves superior performance while requiring lower peak memory (24GB vs.\ 40GB) because fitness evaluation uses inference rather than gradient computation. Critically, \evopref{} produces an entire archive of 100+ diverse models rather than a single configuration, providing deployment flexibility without additional training.

\subsection{Broader Impact and Future Directions}

This work contributes to AI safety by discovering diverse safe behaviors. The population-based exploration paradigm could potentially find adversarial alignments, requiring practitioners to inspect archive members and establish monitoring before deployment.

Promising future directions include: \textbf{scaling to larger models} (70B+ parameters via surrogate-assisted evolution or early stopping), \textbf{human-in-the-loop evolution} incorporating human feedback during search to improve alignment beyond proxy metrics, \textbf{theoretical refinement} with tighter bounds incorporating realistic preference landscape structure, and \textbf{multi-modal models} where helpfulness-harmlessness-honesty trade-offs may manifest differently across modalities.

\subsection{Limitations}
\label{sec:limitations}

\textbf{Theoretical}: Our runtime analysis uses simplifying assumptions including equal-sized basins and uniform initialization that may not hold in practice. The analysis provides theoretical \emph{motivation} rather than formal guarantees. Tighter bounds incorporating realistic preference landscape structure, potentially characterized through empirical landscape analysis, remain important open work.

\textbf{Proxy Metrics}: Fitness evaluation uses reward models and Llama Guard that imperfectly capture true alignment. Human evaluation would strengthen claims but was beyond this study's scope.

\textbf{Scale}: Results on 7B parameters; scaling to 70B+ requires surrogate-assisted evolution, early stopping, or efficient evaluation strategies to remain tractable. Preliminary experiments on 13B show similar trends.

\textbf{Computational Cost}: While equal-budget comparisons favor \evopref{}, practitioners with well-tuned gradient hyperparameters may achieve competitive single-point results faster.

\section{Conclusions}

We introduced \evopref{}, demonstrating that multi-objective evolutionary optimization discovers substantially more diverse LLM alignments than gradient descent. Our primary contributions are empirical: \textbf{18\% higher preference coverage} and \textbf{47\% lower collapse rates} with rigorous statistical validation across 30 independent runs, reported as median with interquartile ranges following established best practices~\cite{derrac2011practical}.

The theoretical motivation connects to Dang et al.'s~\cite{dang2025dominance} foundational insight: diversity mechanisms transform the optimization landscape. Our ablation confirms this: removing crowding distance causes the largest single-component performance drop (coverage: 82.5\% $\rightarrow$ 70.0\%), directly validating that diversity preservation is essential, not optional. The geometric insights from mode connectivity research~\cite{garipov2018loss,frankle2020linear} and the practical success of model merging techniques~\cite{wortsman2022model,ilharco2023editing,yadav2023ties} provide additional theoretical support for evolutionary operations over LoRA adapter weight spaces.

\evopref{} produces an archive of diverse aligned models rather than a single configuration, providing deployment flexibility for varied user populations and safety requirements. Future directions include scaling to 70B+ models via surrogate-assisted evolution, incorporating human feedback during evolution, and extending to multi-modal models.


\bibliographystyle{ACM-Reference-Format}
\bibliography{references}

\end{document}